\documentclass[runningheads]{llncs}
\usepackage{graphicx}
\usepackage{tabularx}
\usepackage{mathtools}

\begin{document}

\title{Prostate motion modelling using biomechanically-trained deep neural networks on unstructured nodes}

\author{Shaheer U. Saeed\inst{1} \and
Zeike A. Taylor \inst{2} \and
Mark A. Pinnock \inst{1} \and
Mark Emberton \inst{3} \and
Dean C. Barratt \inst{1} \and
Yipeng Hu \inst{1}}


\authorrunning{Saeed et al.}

\titlerunning{Biomechanically-trained neural networks on unstructured nodes}

\institute{Centre for Medical Image Computing and Wellcome/EPSRC Centre for Interventional \& Surgical Sciences, University College London, London, UK \\  
\and
CISTIB Centre for Computational Imaging and Simulation Technologies in Biomedicine, Institute of Medical and Biological Engineering, University of Leeds, Leeds, UK \\  
\and
Department of Urology, University College London Hospitals NHS Foundation Trust, London, UK \\
\email{zcemsus@ucl.ac.uk}  
}

\maketitle

\begin{abstract}

In this paper, we propose to train deep neural networks with biomechanical simulations, to predict the prostate motion encountered during ultrasound-guided interventions. In this application, unstructured points are sampled from segmented pre-operative MR images to represent the anatomical regions of interest. The point sets are then assigned with point-specific material properties and displacement loads, forming the un-ordered input feature vectors. An adapted PointNet can be trained to predict the nodal displacements, using finite element (FE) simulations as ground-truth data. Furthermore, a versatile bootstrap aggregating mechanism is validated to accommodate the variable number of feature vectors due to different patient geometries, comprised of a training-time bootstrap sampling and a model averaging inference. This results in a fast and accurate approximation to the FE solutions without requiring subject-specific solid meshing. Based on 160,000 nonlinear FE simulations on clinical imaging data from 320 patients, we demonstrate that the trained networks generalise to unstructured point sets sampled directly from holdout patient segmentation, yielding a near real-time inference and an expected error of 0.017 mm in predicted nodal displacement.

\end{abstract}

\section{Introduction and related work}

Computational biomechanical modelling has applications in the areas of computer aided intervention, surgical simulation and other medical image computing tasks \cite{Hu_data_fusion,Hu_MR_to_US,Haou_AR,Erhart_FEA_aneur}. In particular, numerical approaches based on finite element (FE) analysis have been applied in a wide range of clinical tasks, such as modelling soft tissue deformation in augmented reality \cite{Haou_AR} and medical image registration \cite{Hu_MR_to_US,Bharatha_prostate_reg}. For example, during transrectal ultrasound (TRUS) guided prostate biopsy and focal therapy, FE-based biomechanical simulations have been proposed to predict physically plausible motion estimations for constraining the multimodal image registration \cite{Hu_MR_to_US,Hu_data_fusion,Wang_MR}.

Due to the highly nonlinear behaviour of soft tissues, complex anatomical geometry and boundary condition estimates, FE simulations often rely on iterative algorithms that are computationally demanding. Many developments have made the modern FE solver highly efficient, such as using parallel computing algorithms with graphics processing units (GPUs) \cite{Taylor_GPU}, simplifying the mechanical formulation \cite{Lee_FEA_simple1,Cotin_FEA_simple2}, or learning reduced-order solutions \cite{Taylor_red_order}. However, a real-time solution remains challenging. In one of the approaches for the prostate imaging application, an average meshing-excluded computation time of 16s per simulation, on a GPU, was reported \cite{Hu_MR_to_US}. To meet the surgical requirement in efficiency, many turned to statistical learning approaches that summarise the FE simulations with a lower dimensional approximation \cite{Hu_smm,Saito_ssm,Khallaghi_stat_reg,Wang_MR}.

Largely motivated by the representation capability and the fast inference, deep neural networks have been used to reduce computation time for biomechanical modelling problems. Meister et al have proposed to use neural networks to approximate the time integration, which allowed much larger time steps to accelerate the iterative optimisation \cite{Meister_DL}. Mendizbal et al presented a methodology for estimating the deformations from FE simulations using a U-Net variant, to efficiently predict a deformation field on regularly sampled grid nodes \cite{Mendi_hy}. U-Mesh was able to make approximations for deformations in the human liver under an applied force and a mean absolute error of 0.22 mm with a prediction time of 3 ms was reported \cite{Mendi_hy}. The model requires, as input, point clouds derived from tetrahedral meshes mapped to a sparse hexahedral grid. It also assumed that the deformable region has uniform material properties throughout. Liang et al used deep learning to estimate stress distributions by learning a mapping from a geometry encoding to stress distribution \cite{Liang_FEA_aorta}. The model was trained on FE simulation data and was able to predict stress distributions in the aortic wall given an aorta geometry \cite{Liang_FEA_aorta}. The constitutive properties were also assumed invariant throughout the entire geometry \cite{Liang_FEA_aorta}.

In this work, we adapt a PointNet \cite{Pointnet} with a bootstrap aggregating sampling strategy, to model the biomechanics on a set of input feature vectors that represent patient-specific anatomy with node-wise boundary conditions and material properties. The use of unstructured data to represent the geometry potentially alleviates the need for meshing or shape encoding via principal component analysis. The incorporation of material properties within the proposed input feature vectors allows accommodation of more realistic inhomogeneous biological materials. We integrated these changes into the proposed PointNet based neural network training for deformation prediction tasks. The PointNet has been applied for a wide range of learning tasks with un-ordered point clouds as input, such as classification and segmentation. 

We summarise the contributions in this work: 1) The proposed network provides a permutation-invariant deep network architecture over the unstructured input feature vectors, additionally allowing flexible point set sampling schemes without requiring pre-defined number of points or spatially regular nodal locations; 2) Out-of-nodal-sample generalisation ability is also investigated, based on the input feature vectors directly sampled from segmentation on holdout patient data; 3) The efficiency and accuracy of the proposed method is validated using a large holdout patient data set (30,000 simulations from 60 patients) in a real clinical application for predicting TRUS probe induced deformations in the prostate gland and the surrounding regions.

\section{Methods}

In Sections 2.1-2.3, a deep neural network and its training strategy are described to learn a high-dimensional nonlinear mapping between a set of input feature vectors representing the geometry, applied loads and material properties, and a set of displacements on the input nodal locations.

\subsection{Unstructured input feature vectors}

Without loss of generality, let $\{\mathbf{x}_n\}$ be a set of $N$ un-ordered feature vectors ${\mathbf{x}}_n=[{\mathbf{p}}_n^T,{\mathbf{b}}_n^T, {\mathbf{k}}_n^T]^T$, where $n=1,2,...,N$. $\{\mathbf{p}_n\}$ are the point coordinates; $\{\mathbf{b}_n\}$ represent the externally applied loads with known boundary conditions; and $\{\mathbf{k}_n\}$ specify the parameter values of the material property models. 

For the prostate motion modelling application, $\mathbf{p}_n=[x_n, y_n, z_n]^T$ contain 3D Euclidean coordinates of the sampled points. $\mathbf{k}_n=[G_n,K_n]^T$ contain the shear and bulk moduli in an isotropic, elastic neo-Hookean material model \cite{FEA_2}. The nodes with available displacement loads are assigned with vectors $\mathbf{b}_n=[1, b^x_n, b^y_n, b^z_n]^T$ where $1$ indicates the availability of the assigned displacement for the node, while those without are assigned $\mathbf{b}_n=[0,0,0,0]^T$. This representation can be readily generalised to dimension-specific loads by adding more 'switches', i.e. the first elements of the current vectors, and to other types of boundary conditions such as force.

\begin{figure}[h!]
\includegraphics[width=\textwidth]{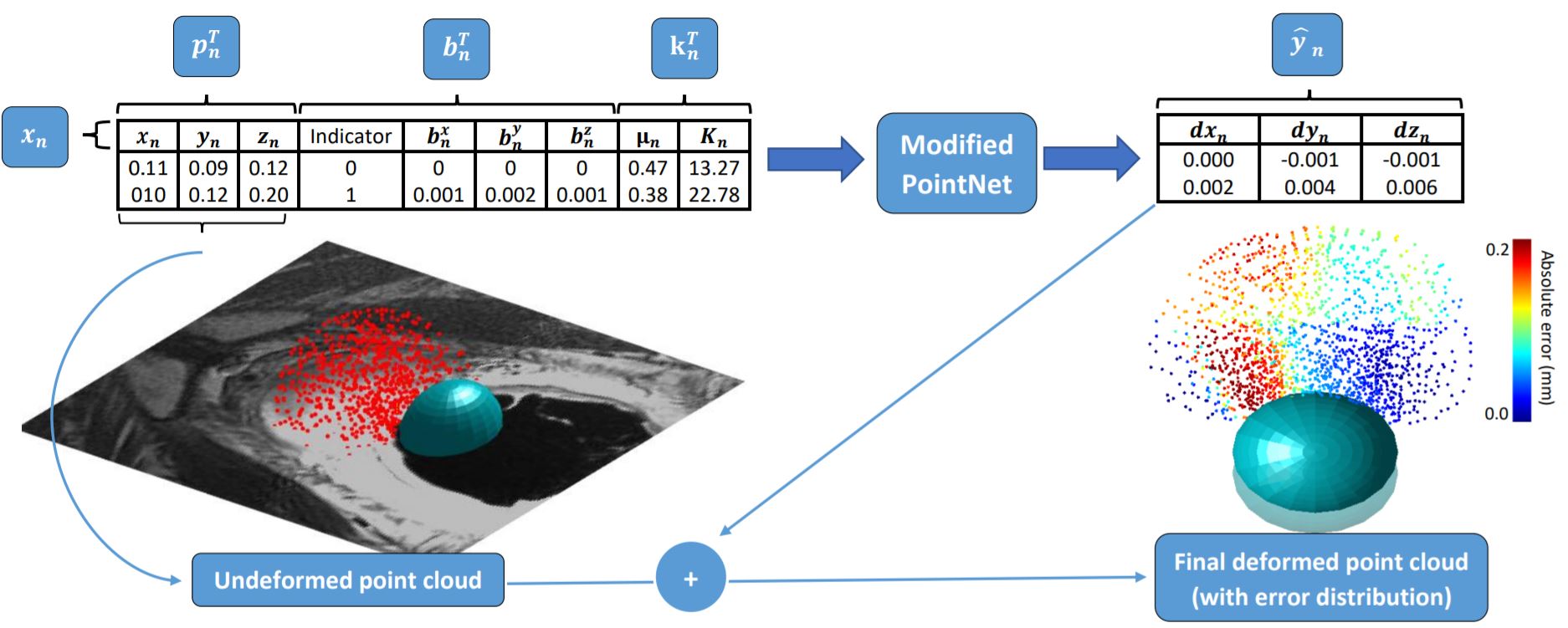}
\caption{Point cloud of prostate (red) with data flow (blue arrows) and error distribution of deformation prediction due to simulated TRUS probe (green sphere) movement.}
\label{Viz}
\end{figure}

\begin{figure}[h!]
\includegraphics[width=\textwidth]{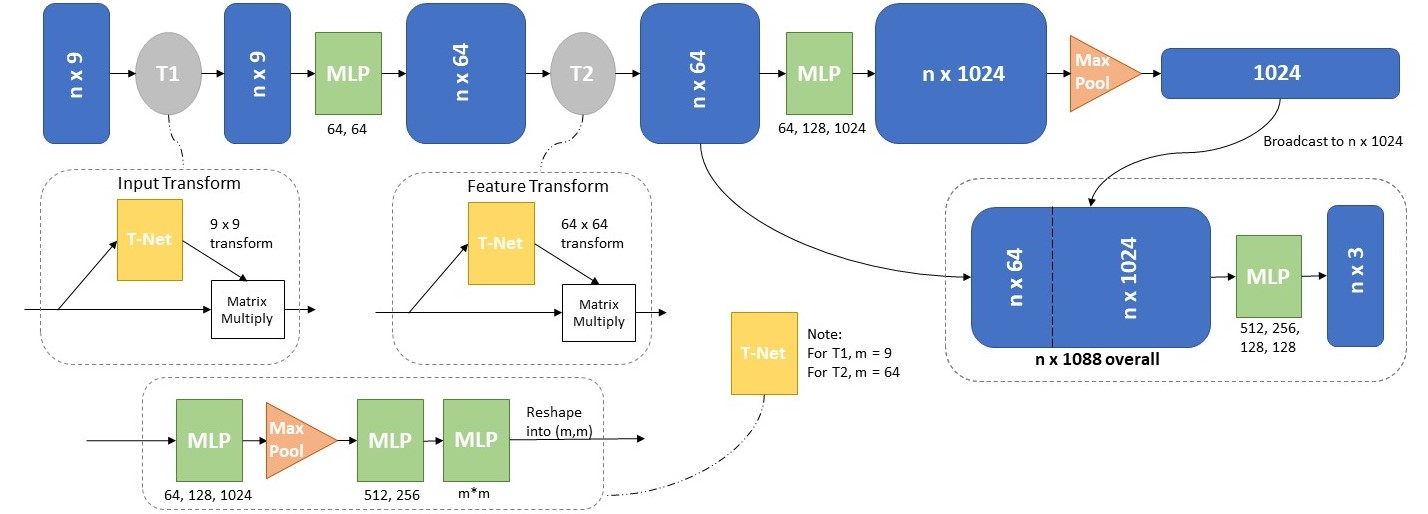}
\caption{Adapted PointNet architecture used for the displacement prediction task. Solid lines show data flow, blue boxes show data size and dashed lines show expanded views.} 
\label{PointNet}
\end{figure}

\subsection{Permutation-invariant nodal displacement prediction}

The PointNet is adapted to predict displacement vectors $\{\mathbf{\hat{y}}_n\}$ from input features $\{\mathbf{x}_n\}$, as illustrated in Fig.~\ref{Viz}. The adapted PointNet architecture is illustrated in Fig.~\ref{PointNet}, generalising the first transformation net T1-net to 9D space instead of the original 3D space. The readers are referred to \cite{Pointnet} for other architectural details.

In this work, ground-truth displacements $\{\mathbf{y}_n\}$ are computed from finite element simulations (see Sections 3.1 and 3.2). Mean squared error is minimised as the loss function to optimise the network weights: $\mathcal{L}(y , \hat{y}) = \Sigma_{n=1}^N (\mathbf{y}_n - \mathbf{\hat{y}}_n)^2/N$.

\subsection{Training-time bootstrap aggregating}

Although the adapted PointNet in theory accepts variable numbers of feature vectors as input during training and inference, an implementation with a fixed number of input vectors is much more efficient \cite{Pointnet} with modern GPU parallel computing support. Reliably and efficiently mapping the often irregularly-shaped input to regular space, such as a cubic grid or a fixed number of points, remains an interesting challenge. 

We propose an alternative bootstrap aggregating training strategy to randomly sample, with replacement, the input feature vectors, in each optimisation iteration. During inference, the final prediction is computed by averaging the predictions from a number of forward-passes of the trained network, which cover all the input feature vectors. We note that the expected back-propagated gradient remains the same using the proposed sampling and averaging scheme, as when training using all the input feature vectors. The proposed sampling-averaging scheme provides a flexible mechanism for training and prediction with patient anatomy represented by different point sets, without restriction on number of points sampled from varying patient geometry. The bootstrap aggregating, also known as bagging, is a model averaging strategy that may also improve the generalisability of the trained network \cite{ESL}.

In this work, the tetrahedron mesh nodes are used to train the network (Section 3.2), after solid meshing of the patient geometry. All mesh nodes have an equal probability of being sampled during each training iteration. This results in a relatively sparse representation of the patient geometry, during the bootstrap aggregating training. The same sampling and model averaging strategy are also applicable to input feature vectors with different and, potentially, simpler point sampling schemes other than finite element mesh nodes. One such example is validated in this study and described in Section 3.4.

\section{Experiments}

All the patients are randomly sampled into three groups, training, validation and holdout sets. For all experiments, we compute mean absolute error (MAE), and the standard deviation (St.D.), between FE simulation results and network predictions. Specific for this application, accurate displacement prediction on the prostate gland and its zonal structures is of importance in guiding both targeted biopsy and focal therapy procedures. Therefore, nodal displacement errors are also computed for each of the gland regions, represented by central zone (CZ) and whole gland (WG) with transition- and peripheral zones. Additionally, the first, second and third quartiles (Q1, Q2 and Q3 respectively) are reported for all nodes. Paired T-test results are reported for holdout set experiments.

\subsection{Data acquisition and finite element simulations}

T2-weighted MR images were acquired from 320 prostate cancer patients who underwent TRUS-guided transperineal targeted biopsy or focal therapy. Tetrahedron meshes of approximately $0.2 \times 0.2 \times 0.2 m^3$ volume of the patient abdominal region were generated, after the prostate glands and surrounding structures were manually segmented. In each simulation, two types of boundary conditions were applied: zero displacement on pelvic bones and sampled nodal displacements on the simulated TRUS probe. Sampled material properties were assigned to elements in different anatomical regions and zonal structures. For each patient, 500 finite element (FE) simulations were performed using NiftySim \cite{Taylor_GPU2}, resulting in 160,000 simulated motion data. Each predicts one plausible prostate motion due to change of ultrasound probe movement, acoustic balloon dilation, bony constraints and varying material properties. Averaged over all simulations, the maximum displacement value in the ground truth data is 5.21 mm, and the mean displacement over the entire point cloud is 0.83 mm. Further details of the finite element modelling can be found in \cite{Hu_MR_to_US,Hu_data_fusion}. Similar FE simulation strategies have been widely used and validated in the same applications \cite{Bharatha_prostate_reg,Alterovitz_reg,Crouch_reg,Bois_reg,Wang_MR}.

\subsection{Network training and hyperparameter tuning}

The point cloud node locations, boundary conditions and material properties used in the FE simulations were assembled into the input feature vectors $\{\mathbf{x}_n\}$, as the PointNet input. Batch normalization was used at each hidden layer, with additional dropout regularisation at a drop rate of 0.25, on the final fully connected layer. The Adam optimiser was used to minimize the mean squared error loss function, with an initial learning rate of 0.001 and a minibatch size of 32. The number of input feature vectors used for training was 14500. Based on preliminary results on the validation set, varying these hyperparameters produced limited impact on the network performance. Conversely, the global feature vector (GFV) size may impact performance significantly \cite{Pointnet}, as it linearly correlates with the number of trainable parameters, and affected model training and inference time in our validation experiments. Therefore, we report the network performance results with different GFV sizes, on the validation set. All the other hyperparameters were configured empirically and remained fixed throughout the validation and holdout experiments presented in this paper. 

The networks were trained using 100,000 FE simulations from 200 patients, while validation and holdout sets each comprised data from 60 patients (30,000 simulations). Each network took approximately 32 hours to train on two Nvidia Tesla V100 GPUs. 

\subsection{Sensitivity analysis on material properties} 

Several previous studies have trained deep neural networks to predict finite element solutions with uniform material properties in the regions of interest \cite{Mendi_hy,Mendi_liv,Liang_FEA_aorta}. This in principle is unrealistic for human anatomy; e.g. the prostate peripheral, transitional, and central zones are generally accepted to have differing material properties. The importance of material heterogeneity depends on the application, though Hu et al.'s results suggest it is significant for prostate motion predictions \cite{Hu_MR_to_US}. The proposed network readily includes material properties in the input feature vectors. We therefore investigated the difference between networks trained with and without region-specific material properties by treating region-specificity as a hyperparameter. Networks with homogeneous materials were correspondingly trained with reduced input feature vectors, which excluded material parameters: ${\mathbf{x}}_n=[\mathbf{p}_n^T,\mathbf{b}_n^T]^T$.

\subsection{Generalisation to points sampled directly from segmentation}

To demonstrate the generalisability of the network to alternative point sampling methods, we re-sampled all the point locations in the holdout set using a well-established region-wise cuboid tessellation approach. Each of the $a$ anatomical regions was represented by $c$ cuboid tessellations computed from segmentation surface points, without using the solid tetrahedral mesh. From each tessellation, $f$  points were then randomly generated, this resulted in $cf$ points per region and $acf$ points in total. For training and validation purposes, material properties were assigned based on which tessellation the node belongs to, while BCs were interpolated using an inverse Euclidean-distance transform based on five nearest neighbouring nodes. It is noteworthy that, in general, $acf \neq n$ and the bootstrap aggregating, described in Sect. 2.3, still applies to the new input feature vectors.

\section{Results}

\paragraph{Global feature vector size and material properties}

Summarised in Table~\ref{val_tab}, results on the validation set suggest that increasing GFV size, up to a maximum of 1024, reduced error rates. However, further increasing GFV size from 1024 to 2048, the mean error increased from 0.010$\pm$0.011 to 0.013$\pm$0.015. This small yet significant increase (p-value$<$1e-3) may suggest overfitting due to larger network size. Therefore, a GFV size of 1024 was subsequently used in holdout experiments. Also shown in Table~\ref{val_tab}, excluding material properties from the network inputs significantly increased nodal displacement errors, from 0.01$\pm$0.011 to 0.027$\pm$0.029 (p-value$<$1e-3). Material parameters were therefore retained in the input feature vectors in the reported results based on the holdout set.

\paragraph{Network performance on holdout set}

As summarised in Table~\ref{val_tab}, the overall MAEs, on the holdout set, were 0.010$\pm$0.012 mm and 0.017$\pm$0.015 mm, for the FE tetrahedral mesh nodes and the points from the tessellation-sampling, respectively, with a significant difference (p-value$<$1e-3). However, when these network-predicted displacements were compared with the nodal displacements produced with FE simulations, there was no significance found with p-values of 0.093 and 0.081, respectively. The error distributions were skewed towards zero, as can be observed based on the median, $25^{th}$ and $75^{th}$ percentile values reported in Table~\ref{val_tab}. Computed over all the cases in the holdout set, the average MAEs were 0.34 mm and 0.48 mm, for the points sampled from FE mesh nodes and points sampled using tessellation, respectively. We also report an inference time of 520 ms, when predicting displacements for approximately 14,500 nodes, using one single Nvidia GeForce 1050Ti GPU.

\begin{table}[h]
\caption{Results from different experiments. All results presented in mm.}
\label{val_tab}
\begin{tabular}{|p{2.75cm}|p{1.85cm}|p{1.0cm}|p{1.0cm}|p{1.0cm}|p{1.85cm}|p{1.85cm}|}
\cline{2-7}
\multicolumn{1}{c}{} & \multicolumn{4}{|c|}{All points/ nodes} & \multicolumn{1}{c}{CZ}& \multicolumn{1}{|c|}{WG}\\
\cline{2-7}
\multicolumn{1}{c|}{} &   MAE$\pm$St.D.   &   Q1  &   Q2  &   Q3  &MAE$\pm$St.D.  &MAE$\pm$St.D.   \\
\hline
GFV Sizes & \multicolumn{6}{c|}{Results on validation set}\\
\hline
256 &0.113$\pm$0.091      &0.043      &0.094      &0.165      &0.107$\pm$0.080      &0.110$\pm$0.094    \\
\hline
512 &0.065$\pm$0.047      &0.002      &0.031      &0.086      &0.058$\pm$0.052      &0.063$\pm$0.049    \\
\hline
1024&0.010$\pm$0.011      &0.000      &0.005      &0.013      &0.008$\pm$0.013      &0.009$\pm$0.012    \\
\hline
2048&0.013$\pm$0.015      &0.000      &0.006      &0.016      &0.009$\pm$0.011      &0.009$\pm$0.016    \\
\hline
Input Feat. Vectors & \multicolumn{6}{c|}{Results on validation set}\\
\hline
${\mathbf{x}}_n=[{\mathbf{p}}_n^T,{\mathbf{b}}_n^T,{\mathbf{k}}_n^T]^T$ &0.010$\pm$0.011      &0.000      &0.005      &0.013      &0.008$\pm$0.013      &0.009$\pm$0.012     \\
\hline
${\mathbf{x}}_n=[{\mathbf{p}}_n^T,{\mathbf{b}}_n^T]^T$  &0.027$\pm$0.029      &0.000      &0.009      &0.031      &0.023$\pm$0.021      &0.029$\pm$0.020     \\
\hline
Sampling Strategy & \multicolumn{6}{c|}{Results on holdout set}\\
\hline
Tetrahedral Mesh         &0.010$\pm$0.012      &0.000      &0.007      &0.018      &0.009$\pm$0.013      &0.010$\pm$0.012    \\
\hline
Tessellation   &0.017$\pm$0.015      &0.000      &0.009      &0.019      &0.014$\pm$0.020      &0.017$\pm$0.021    \\
\hline
\end{tabular}
\end{table}

\section{Discussion}

Based on statistical test results on the holdout set, reported in Section 4, we conclude that the models presented in this study can predict highly accurate nodal displacements for new patients in our application. This provides an efficient alternative to FE simulations in time critical tasks, where additional computation can be offloaded to the model training phase, such as surgical simulation \cite{Berkley_surg_sim} and surgical augmented reality \cite{Haou_AR} where computation times need to be curtailed during deformation prediction. For the MR-to-TRUS registration application described in Section 1, rather than replacing the FE simulations for constructing a patient-specific motion model for intervention planning, we conjecture a possibility to optimise the deformation prediction directly during the procedure, enabled by the highly efficient inference with non-iterative forward network evaluation.

An important contribution in this work is to provide evidence that, the proposed network can generalise to input feature vectors sampled from alternative methods other than the FE mesh, a simple and versatile tessellation method in this case. This is an important avenue for further investigation of the proposed method, especially for point spatial locations integrated with features like other types of BCs, material parameters and even other physiological measurements of clinical application interests.

The computational cost of data generation and model training is significant and this can be a hindrance in the deployment of deep learning models in practice. Augmentation can serve to alleviate a part of the problem by presenting augmented examples to the network for training. Re-sampling, like regional tessellation is a viable candidate, although this has yet to be investigated. More interestingly, the model generalisability to cases other than probe induced deformation prediction has not been investigated and this presents an opportunity for further research to investigate training strategies such as transfer learning and domain adaptation, which may substantially reduce the computational cost.

\section{Conclusions}
We have presented a PointNet based deep neural network learning biomechanical modelling from FE simulations, for the case of TRUS probe induced deformation of the prostate gland and surrounding regions. The PointNet architecture used for this task allows for randomly sampled unstructured point cloud data, representing varying FE loads, patient anatomy and point-specific material properties. The presented approach can approximate FE simulation with a mean absolute error of 0.017 mm with a sub-second inference. The method can be generalised to new patients and to randomly sampled 3D point clouds without requiring quality solid meshing. The proposed methodology is applicable for a wide range of time critical tasks, and we have demonstrated its accuracy and efficiency with clinical data for a well-established prostate intervention application.

\section*{Acknowledgments}
This work is supported by the Wellcome/EPSRC Centre for Interventional and Surgical Sciences (203145Z/16/Z). ZAT acknowledges funding from CRUK RadNet Leeds Centre of Excellence (C19942/A28832).

\end{document}